\documentclass{tlp}

\usepackage{amsmath,amssymb}
\usepackage{graphicx}
\usepackage{multirow}
\usepackage{hyperref}
\usepackage{proof-at-the-end}
\usepackage{comment}
\usepackage{tikz}
\usepackage{needspace}
\usepackage{multicol}
\usetikzlibrary{decorations.markings}

\long\def\comment#1{}
\usepackage[linesnumbered,ruled,vlined]{algorithm2e}

\begin{document}

\newtheorem{proposition}{Proposition}
\newtheorem{definition}{Definition}
\newtheorem{theorem}{Theorem}
\newtheorem{conjecture}{Conjecture}
\newtheorem{lemma}{Lemma}
\newtheorem{note}{Note}
\newtheorem{corollary}{Corollary}
\newtheorem{example}{Example}

\newcommand{\Line}[1]{line \ref{#1}}
\newcommand{\Theorem}[1]{Theorem~\ref{#1}}
\newcommand{\Corollary}[1]{Corollary~\ref{#1}}
\newcommand{\Definition}[1]{Definition~\ref{#1}}
\newcommand{\Lemma}[1]{Lemma~\ref{#1}}
\newcommand{\Proposition}[1]{Proposition~\ref{#1}}
\newcommand{\Example}[1]{Example~\ref{#1}}
\newcommand{\Section}[1]{Section~\ref{#1}}

\newcommand{\K}{\mathcal{K}}
\newcommand{\pro}{\mathcal{P}}

\newcommand{\Tmod}{$\mathds{T}$-stable model}
\newcommand{\emodels}{\models_{E}}
\newcommand{\mmodels}{\models_{\text{MKNF}}}
\newcommand{\nm}{\not\models}
\newcommand{\nmm}{\not\mmodels}
\newcommand{\ont}{\mathcal{O}}
\newcommand{\ka}{$\mathbf{K}$-atom}
\newcommand{\KA}{\mathbf{KA} (\mathcal{K})}
\newcommand{\KAN}{\mathbf{KA}^- (\mathcal{K})}
\newcommand{\KAO}{\mathbf{KA} (\mathcal{\ont})}
\newcommand{\KAON}{\mathbf{KA}^- (\mathcal{\ont})}

\newcommand{\KAH}{Head}
\newcommand{\known}{\mathbf{K}}
\newcommand{\taught}{\mathbf{t}}
\newcommand{\contra}{\mathbf{f}}
\newcommand{\no}{\mathbf{not}}
\newcommand{\ki}{$\mathbf{K}$-interpretation}
\newcommand{\hti}{$HT$-interpretation}
\newcommand{\kpo}{\known\pi(\ont)}
\newcommand{\po}{\pi(\ont)}
\newcommand{\pk}{\pi(\K)}
\newcommand{\pp}{\pi(\pro)}
\newcommand{\pr}{\pi(r)}
\newcommand{\comprule}{\pro_{rule}}
\newcommand{\compsatr}{\ont_{satr}(\hi)}
\newcommand{\compsup}{\K_{sup}(\hi)}
\newcommand{\osigma}{\overline{\sigma}}
\newcommand{\rh}{head(r)}
\newcommand{\rb}{body(r)}
\newcommand{\prb}{body^+(r)}
\newcommand{\nrb}{body^-(r)}
\newcommand{\rbf}{Body(r)}
\newcommand{\T}{\mathbf{T}}
\newcommand{\F}{\mathbf{F}}
\newcommand{\epsi}{\mathcal{E}}
\newcommand{\hi}{\hat{I}}
\newcommand{\gk}{G(\K)}
\newcommand{\go}{G(\ont)}
\newcommand{\gp}{G(\pro)}
\newcommand{\ia}{\hi \setminus \{a\}}
\newcommand{\ilset}{\hi \setminus L}
\newcommand{\ilsetp}{(\hi \setminus L)}
\newcommand{\diffset}{\hi \setminus \hi'}
\newcommand{\diffsetp}{(\hi \setminus \hi')}
\newcommand{\lk}{Loops(\K)}
\newcommand{\varcomp}{var(\Delta_{\K})}
\newcommand{\AI}{A^{\hi}_{\K}}
\newcommand{\gop}{\gamma_{\ont}^+}
\newcommand{\gon}{\gamma_{\ont}^-}
\newcommand{\dk}{\Delta_{\K}}
\newcommand{\lnk}{\Lambda_{\K}}
\newcommand{\AIvar}{({\AI}^T \cup {\AI}^F)}
\newcommand{\Kintro}{\K = (\pro, \ont)}
\newcommand{\kli}{\K_{loop} (\hi)}
\newcommand{\kci}{\K_{comp}(\hi)}
\newcommand{\mknfhead}{\known h_0\wedge\cdots\wedge\known h_i}
\newcommand{\mknfprb}{\known p_0\wedge\cdots\wedge\known p_j}
\newcommand{\mknfnrb}{\no n_0 \wedge ... \wedge \no n_k}
\newcommand{\mknfrb}{\known p_0\wedge\cdots\wedge\known p_j \wedge \no n_0 \wedge ... \wedge \no n_k}
\newcommand{\shorthands}{$\rh = \{h_0\wedge\cdots\wedge h_i\}$, $\prb = \{p_0\wedge\cdots\wedge p_j\}$,
and $\nrb = \{n_0\wedge\cdots\wedge n_k\}$}

\lefttitle{Riley Kinahan, Spencer Killen, Kevin Wan, and Jia-Huai You}
\jnlPage{1}{x}
\jnlDoiYr{2024}
\doival{10.1017/xxxxx}

\title[Foundations of Conflict Driven Solving For Hybrid MKNF]{On the Foundations of Conflict-Driven Solving for Hybrid MKNF Knowledge Bases}

\begin{authgrp}
\author{\gn{Riley} \sn{Kinahan}, \gn{Spencer} \sn{Killen}, \\ \gn{Kevin} \sn{Wan}, and \gn{Jia-Huai} \sn{You}}
\affiliation{University of Alberta, Edmonton, Alberta, Canada \\ 
\emails{rdkinaha@ualberta.ca}{sjkillen@ualberta.ca}{kcwan1@ualberta.ca}{jyou@ualberta.ca}}

\end{authgrp}

\history{\sub{05 08 2024;} \rev{07 15 2024;} \acc{xx xx xxxx}}

\maketitle

\begin{abstract}
Hybrid MKNF Knowledge Bases (HMKNF-KBs) 
constitute a formalism for tightly integrated reasoning
over closed-world rules and open-world ontologies. 
This approach allows for accurate modeling  
of real-world systems,
which often rely 
on both categorical and normative reasoning.
Conflict-driven solving 
is the leading approach for computationally hard problems, 
such as satisfiability (SAT) and answer set programming (ASP), 
in which MKNF is rooted.
This paper
investigates the theoretical underpinnings
required for a conflict-driven solver of HMKNF-KBs.
The approach defines a set of completion and loop formulas, 
whose satisfaction characterizes MKNF models.
This forms the basis for a set of nogoods, which in turn
can be used as the backbone for a conflict-driven solver.
\end{abstract}


\section{Introduction}
Real-world problems
often require integrated reasoning
of both rules-based and ontological knowledge, 
spanning domains such as customs, healthcare, 
and penal systems~\citep{knorr2021combining,alberti2012normative}. 
For example, in customs, 
ontological reasoning aids in categorizing imported goods, 
while rule-based reasoning determines inspection procedures~\citep{knorr2021combining}.
The prevalence of such applications
has led to the development of frameworks
for reconciling ontologies with rules, 
including Hybrid MKNF Knowledge Bases~\citep{reconciling_dls_and_rules}. 

A Hybrid MKNF Knowledge Base (HMKNF-KB) consists of two components: 
a logic program of rules, such as in answer set programming (ASP),
and an ontology representable 
in a decidable fragment of first-order logic, 
most often under a description logic.
The main feature of HMKNF-KBs, 
compared to other approaches that combine ASP with description logics,
is the {\em tight} integration between their two components.
Here, tightness refers to the ability of an integration
to allow for derivation within one component based on conclusions from the other.
For certain applications, a one-way flow of information is sufficient.
However, greater tightness results in a richer interplay between two knowledge sources.

Some integrations have partial tightness, such as {\em dl-programs} \citep{eiter2005nonmonotonic}.
While these allow for back-and-forth derivation between two components, it must be localized to specific dl-atoms within rules, which may query the ontology.
Conversely, HMKNF-KBs fully integrate the two components \-- inferences in one are immediately available to the other. 
The integration also has several other desirable properties, including 
{\em faithfulness} to each of the underlying components,
{\em flexibility} in whether any predicate can be viewed under  closed or open-world reasoning,
and {\em decidability} (under the \emph{DL-safety} assumption)
\citep{reconciling_dls_and_rules}.\footnote{ 
    There are other systems of tight integration of rules and first-order formulas
    such as IDP-systems \cite{idp2008}, however they are beyond the scope of
    this paper since their reasoning task deals with extension of classical
    first-order logic which presents a very different challenge.
}

The well-founded semantics for Hybrid MKNF~\citep{knorr2011local}
have enjoyed considerable focus
due to their polynomial complexity (under some assumptions), 
and form the basis for the reasoner NoHR~\citep{kasalica2020nohr}.
However, reasoning tools for the stable model semantics
are still relatively limited. 
\cite{ji2017well} define well-founded operators for non-disjunctive Hybrid MKNF.
\cite{killen2021} generalize this approach to the disjunctive case
for a DPLL-based solver~\citep{dpll}.
This represents significant progress towards efficient solving, 
but is still behind state-of-the-art conflict-driven solving, 
which is widely adopted by ASP and its extensions.

While there have been advancements in resolution or 
query-based solving for Hybrid MKNF
\citep{alferes2013}, 
we strictly focus on solvers that employ bottom-up model-search.
The prominent modern ASP solvers, Clasp~\citep{gebser2013advanced} 
and WASP~\citep{alviano2015advances}
(\comment{developed }within the Potassco/Clingo~\citep{gebser2019multi} and DLV systems~\citep{adrian2018asp} resp.),
combine conflict-driven SAT solving
with native ASP propagation,
to achieve a high degree of performance.
Clingo is designed to be extensible, 
and numerous applications have been built on it. 
The relevant case here
is the system DLVHEX~\citep{redl2016dlvhex,eiter2018dlvhex}, 
which allows integrating rules with arbitrary external sources
and can solve dl-programs~\citep{eiter2006towards}. 
Despite being only partially tight, dl-programs are closely related to HMKNF-KBs. 
It has been shown 
that under reasonable assumptions,
HMKNF-KBs can be translated to dl-programs~\citep{eiter2015linking}.
Thus, one can translate HMKNF-KBs to dl-programs
and use DLVHEX to compute  the MKNF models;
however, 
it is unclear whether this would be as powerful 
as a native conflict-driven approach.

Therefore, this paper aims
to develop a general theory
relating HMKNF-KB 
model computation
to conflict-driven solving. 
In particular, we focus on the notion of a {\em \ki},~which 
represents everything concludable under an MKNF model.
To accomplish this task, 
we first characterize whether \ki s correspond to models
according to whether they satisfy a set of formulas;
we call these the \emph{completion} and \emph{loop formulas}.
They follow the naming convention of,
and are directly inspired by,
the formulas of \citep{loop_formulas},
which characterize answer sets of disjunctive logic programs.
We then define \emph{nogoods} in the sense of \citep{gebser2012}, 
which capture the constraints induced by our formulas
and thereby characterize models of HMKNF-KBs.
Finally, 
we give an overview of how our nogoods can be used
within conflict-driven algorithms.
We conclude with comments on related and future work, 
including practical considerations for implementing a solver.

Complete proofs are provided in a separately available Appendix. 
To aid reading, we provide
proof sketches for the two most crucial claims.

\section{Preliminaries}

\noindent
{\bf Minimal Knowledge and Negation as Failure (MKNF):}

MKNF is a nonmonotonic logic formulated by Lifschitz~\citeyearpar{lifschitznonmonotonic1991}. 
MKNF formulas extend first-order formulas with two modal operators,
$\known$ for minimal knowledge, 
and $\no$ for negation as failure.
We define a first-order interpretation $I$ as usual and denote the universe of $I$ by $|I|$.  
An {\em MKNF structure} is a triple $(I,M,N)$, 
where $M$ and $N$ are sets of first-order interpretations
within the universe $|I|$. 
The language of MKNF formulas contains a constant for each element of $\left|I\right|$, which
we call a {\em name}.
We define the satisfaction relation between an MKNF structure $(I,M,N)$ and an MKNF formula as follows:

\vspace{-.1in}
\begin{align*}
    &(I,M,N) \models \phi \text{ ($\phi$ is a first-order atom)  if } \phi \text{ is true in } I, \\
    &(I,M,N) \models \neg \phi \text{ if } (I,M,N) \not \models \phi, \\
    &(I,M,N) \models \phi_1 \wedge \phi_2 \text{ if } (I,M,N) \models \phi_1 \text{ and } (I,M,N) \models \phi_2, \\
    &(I,M,N) \models \exists x \phi \text{ if } (I,M,N) \models \phi[a\setminus x] \text{ for some } a, \\
    &(I,M,N) \models \known \phi \text{ if } (J,M,N) \models \phi \text{ for all } J\in M, \\
    &(I,M,N) \models \no\, \phi \text{ if } (J,M,N) \not \models \phi \text{ for some } J \in N. 
\end{align*}


The symbols $\top,\bot,\vee,\forall, \text{ and } \supset$ are interpreted as usual.

An {\em MKNF interpretation} $M$
is a nonempty set of first-order interpretations.
Throughout this work, we employ the {\em standard name assumption} to avoid unintended behaviors \citep{reconciling_dls_and_rules}.
This assumes all interpretations are Herbrand interpretations 
with a countably infinite number of additional constants, 
and that the predicate $\approx$ is a congruence relation.
Thus, we do not explicitly mention the universe associated with interpretations.

An MKNF interpretation $M$ satisfies an MKNF formula $\phi$, written 
$M \mmodels \phi$,
if $(I,M,M) \models \phi$ for each $I \in M$.

\begin{definition}
    An MKNF interpretation $M$
    is an {\em MKNF model} of an MKNF formula $\phi$, 
    if $M \mmodels \phi$, 
    and for all MKNF interpretations $M'$ s.t.\ $M' \supset M$, we have
   $\forall I'\in M, (I',M',M) \nm \phi$.
\end{definition}

\smallskip
\noindent
{\bf Hybrid MKNF Knowledge Bases (HMKNF-KBs):}

\citet{reconciling_dls_and_rules} identify a subset of MKNF formulas as Hybrid MKNF.
In this new language, an HMKNF-KB  $\K = (\pro, \ont)$ consists of
a finite set of {\em rules} termed a \emph{rule base} $\pro$, 
and an {\em ontology} $\ont$ translatable to first-order logic as $\po$.
A rule $r$ is of the form
\begin{align*}
    h_0, \ldots, h_m\leftarrow 
    p_0, \ldots, p_j, \neg
    n_0, \ldots, \neg n_k
\end{align*}
where $h_i$, $p_i$, and $n_i$ are function-free first-order atoms. 
We denote  $\prb=\{p_0,\ldots,p_j\}$, $\nrb=\{n_0,\ldots,n_k\}$, 
$Body(r) = \bigwedge \prb \wedge \neg \bigvee \nrb$,
and $\rh=\{h_0,\ldots,h_m\}$.
A rule $r$'s semantics is governed by the following MKNF formula.
\begin{align*}
    \pi(r)=
    \forall \overrightarrow{x}:
    (\known h_0\vee \cdots \vee\known h_z)\subset
    (\known p_0\wedge\cdots \wedge\known p_j \wedge 
    \no\ n_0 \wedge \cdots \wedge \no\ n_k)
\end{align*}
where $\overrightarrow{x}$ is the vector of free variables in $r$.
Naturally, a rule base $\pro$ translates to an MKNF formula as $\pp = \bigcup_{r \in \pro} \pi(r)$.
We say that an MKNF interpretation $M$
is an {\em MKNF model} of the HMKNF-KB $\K$
if it is an MKNF model of the MKNF formula $\pk = \kpo \wedge \pp$.

A program $\pro$ is ground if no rule $r \in {\pro}$ has variables.
\citet{reconciling_dls_and_rules} show 
that under the assumption of \emph{DL-safety}\footnote{
An HMKNF-KB $\Kintro$ is \emph{DL-safe}, if for all rules $r \in \pro$, 
all variables present in $r$ appear within $\prb$,
under a predicate that does not occur in $\po$.},
a first-order rule base 
is semantically equivalent to a finite ground rule base
and that decidability is guaranteed 
for HMKNF-KBs with decidable ontologies.
In this work, we assume rule bases are ground.

We define $\KA$ be the set
containing every atom $\phi$ that occurs in $\pro$ (either as $\known \phi$ or $\no\ \phi$) and we use $\KAO$ to denote the maximal subset of $\KA$ s.t.\ for each $p(t_1,\ldots,t_m) \in \KAO$, the predicate $p$ also appears in $\po$.

We define the {\em objective knowledge} of an HMKNF-KB $\K$ 
w.r.t.~a set $S \subseteq \KA$ as the set 
$OB_{\ont, S}= \{\po\}\cup\{\phi \ | \ \phi\in S\}$.
Intuitively,
$OB_{\ont, S}$ is a first-order formula that considers $\ont$
with the assumption that $\known \phi$ holds for each $\phi \in S$.
In this paper, we prefer polynomial ontologies, i.e., for any $S \subseteq \KAO$ and $a \in \KAO$,
the relation $OB_{\ont, S} \models a$ can be checked in polynomial time.

The following notion of a \ki\
allows for a simplified representation of an MKNF interpretation
as a single set of atoms.
A central focus of this work is to show which \ki s 
are \emph{induced} by MKNF models.
Such a relationship is not completely straightforward,
due to the quite different nature of the two representations.

\begin{definition}
    A \emph{\ki}\ is any set of atoms $\hi \subseteq \KA$.
    An MKNF interpretation $M$ \emph{induces} a \ki\ $\hi$, 
    if $\hi = \{a \in \KA \ | \ M \mmodels \known a\}$. 
    Whereas, 
    a \ki\ $\hi$ \emph{extends} 
    to an MKNF interpretation $M$ 
    if $M = \{ I \ | \ I \models OB_{\ont, \hi}\}$.
\end{definition}

Above,  we use the notation $\hi$ to express that
as an interpretation in an entailment relation, $\hi \models \phi$,
any atom $a \in \KA$ but not in $\hi$ is assigned to {\em false}.

\comment{
In this paper we explore the relation between \ki s induced by MKNF models
and \ki s which satisfy the formulas defined in \Section{sec: SAT}.
As \ki s are merely sets of atoms we give the following 
definition  clarifying how to treat them as propositional interpretations.
\begin{definition}
    The satisfaction relation between a set of atoms $\hi$ and a formula $\phi$
    is expressed recursively as follows:
    \begin{align*}
        &\text{for an atom } \phi, \hi \models \phi \text{ if } \phi \in \hi, \\
        &\hi \models \neg \phi \text{ if } \hi \nm \phi \\
        &\hi \models (\phi_1 \wedge \phi_2) \text{ if } \hi \models \phi_1 \text{ and }
        \hi \models \phi_2, \\
        &\hi \models (\phi_1 \vee \phi_2) \text{ if }
        \hi \models \phi_1 \text{ or } \hi \models \phi_2 \\
        &\hi \nm \bot \\
    \end{align*}
\end{definition}
}

\comment{
The next definition allows us to more easily discuss 
possible circular derivations.
\begin{definition}
    We say that a set of atoms $S$ {\em externally supports} a set of atoms $A$
    through an ontology $\ont$ if
    $OB_{\ont,S \setminus A} \models  a)$ for each $a \in A$. If $A$ is singleton $\{\phi\}$, we then say $S$ externally supports atom $\phi$.
\end{definition}
}

\medskip \smallskip
\noindent
{\bf Assignments and Nogoods:}

Nogoods \citep{gebser2012} act as canonical representations of Boolean constraints,
reflecting partial \emph{assignments},
which cannot be extended to a \emph{solution}.
A \emph{nogood} $\{\sigma_1, \ldots, \sigma_n\}$,
is a set of literals $\sigma_i$,
of the form $\T v_i$ or $\F v_i$ for $1\leq i \leq n$, 
where $v_i$ is a propositional variable.
The \emph{complement} of a literal,
is referred to by 
$\overline{\T v} = \F v$ and $\overline{\F v} = \T v$.
For any set $\delta$ of literals, 
$\delta^T = \{v \ | \ \T v \in \delta\}$ and 
$\delta^F = \{v \ | \ \F v \in \delta\}$. 
The set of variables occurring within a set of nogoods $\Delta$,
is denoted $var(\Delta)=\bigcup_{\delta \in \Delta} (\delta^\T \cup \delta^\F)$.
An \emph{assignment} $A$ for $\Delta$
is any subset of $\{\T v, \F v \ | \ v \in var(\Delta)\}$
such that $A^\T \cap A^\F = \emptyset$.
A \emph{solution} for $\Delta$
is an assignment $A$ for $\Delta$,
such that $A^\T \cup A^\F = var(\Delta)$ and $\delta \not \subseteq A$ for all $\delta \in \Delta$.
A nogood $\delta$ is \emph{unit-resulting} for an assignment $A$
if $|\delta \setminus A| = 1$.
For a literal $\T \sigma$ in an assignment, $\sigma$ is true within the assignment,
whereas for $\F \sigma$, $\sigma$ is false.

\section{Dependency Graph}
A guiding principle behind the dependency graph of a logic program 
is to provide a syntactic overapproximation
of the true semantic dependency between atoms within the program.
This overapproximation 
can be used to bound the possible sources of circular derivation to \emph{loops}, 
sets of atoms where there is a path between any two in the set. 

Given an HMKNF-KB $\Kintro$, a dependency graph  $G(\K)$
is the graph containing all vertices and edges in either of two dependency graphs, 
$G(\cal P)$ or $G(\cal O)$.
$G(\pro)$ consists of vertices for atoms in $\KA$
and edges from a vertex $a$ to a vertex $b$ if  there is a rule $r \in \pro$ such that $a \in \rh \text{ and } b \in \prb$.
\comment{
    Assume this dependency graph is available.
    Then, 
    the dependency graph $\gk$ is composed of all
    vertices and edges which occur in at least one of $G(\pro)$ or $G(\ont)$. 
}
Below,
we offer a characterization of $\go$ in terms of the entailment relation.
\begin{definition}
\label{dependency-graph}
    An \emph{ontology dependency graph} $G(\ont)$ for an HMKNF-KB $\Kintro$,
    contains vertices for each atom in $\KAO$.
    There is an edge from vertex $a$ to vertex $b$
    in $G(\ont)$ if for some 
    $S \subseteq \KAO$, where $a \not \in S$ and $b \in S$, we have
    \begin{align*}
        &OB_{\ont, S} \models a, 
        &\text{(contributes to derivation)}\\  
        &OB_{\ont, S} \not\models \bot \ \text{and},
		&\text{(is consistent)} \\
        &OB_{\ont, S\setminus \{b\}} \not \models a.
		&\text{(is minimal)}
    \end{align*}
\end{definition}
\comment{\footnote{Other characterizations are possible and this 
is an area where more exploration can be preformed.}}

It is not difficult to see that the ontology dependency graph defined here is an overapproximation
of true semantic dependencies based on entailment relation.  In addition,
there is a unique minimal version of $\go$ containing 
only those edges that are strictly required by its definition.
However we do not expect to tractably generate such a dependency graph in general.
This is because the ontology of an HMKNF-KB is allowed to take a variety of forms,
so dependency graph generation for any particular ontology 
requires separate study.
As such, to keep our apporach general
we assume some version of $\go$ is provided.
To ensure correctness (of overapproximation) 
the assumed dependency graph must simply contain every edge within 
the minimal version, in accordance with Definition 
\ref{dependency-graph}.
A similar approach 
of leveraging externally provided information for dependency pruning
is explored within \citep{eiter2021pruning}, 
to enhance minimality checking.

Similar to~\citep{clark1977negation}, we call 
$\K$ 
\emph{tight} if $G(\K)$ is acyclic. 
We denote by $\lk$
the set of loops of the dependency graph $G(\K)$.

\section{Completion and Loop Formulas}
\label{sec: SAT}
In this section we characterize models of HMKNF-KBs 
through logical formulas.
Our approach follows that of \citet{loop_formulas} who defined \emph{completion} 
and \emph{loop formulas} to capture the answer sets of disjunctive logic programs.
While this work is self-contained, 
we draw frequent comparison to their seminal work to ease understanding.

\subsection{Completion}
\citet{loop_formulas} show that an interpretation of a tight 
disjunctive 
logic program is an answer set
if and only if it satisfies a set of formulas
termed the \emph{completion}.
A program's completion is composed of a \emph{rule completion}, clauses that ensure 
atoms whose truth is implied must be included in an answer set, and a \emph{support completion}, clauses that ensure true atoms within an interpretation
are supported by some rule.

The rule completion of \citep{loop_formulas} can be easily adopted as follows.
\needspace{2 \baselineskip}
\begin{samepage}
\begin{definition}
    The {\em rule completion} of a rule base $\pro$ is $\pro_{rule} = \{  Body(r) \supset \bigvee head(r) 
        \ | \
        r \in \pro \}$.
\end{definition}
\end{samepage}

The formula $\pro_{rule}$ guarantees that 
for any satisfying \ki\ $\hi$,
atoms implied by the rules of $\pro$ under $\hi$ are contained in $\hi$.
We can construct an analogous formula for an ontology $\ont$ that requires all atoms entailed by $\ont$
given $\hi$ to be within $\hi$.
We define this notion as \emph{saturation} and formulate what we call \emph{saturation completion}.
\begin{definition}
    \label{saturated def}
    Given an HMKNF-KB $\Kintro$, 
    a \ki\ $\hi$ of $\K$
    is \emph{saturated} if
    $OB_{\ont,\hi} \nm \bot$,
	and $\forall a \in \KA$ such that $OB_{\ont, \hi} \models a$, $a \in \hi$.
\end{definition}

\begin{definition}
    Let $\Kintro$ be an HMKNF-KB. 
    The {\em saturation completion} of $\K$, with respect to a \ki\ $\hi$, is
    the set $\ont_{satr}(\hi) = \{a \in \KA \cup \{\bot\} \ | \ OB_{\ont, \hi} \models a\}$.
\end{definition}

In the following lemma, we provide alternative characterizations of saturation
to provide further insight into its relevance
and establish a relationship between \ki s and MKNF interpretations.
For instance, characterization \ref{prop:three} below
implies that all \ki s induced by MKNF models are saturated.

\begin{lemma}
    \label{SAT:saturation lemma}
    The following are equivalent for a \ki\ $\hi$.
    \begin{enumerate}
        \item $\hi$ is saturated.
        \item $\hi$ extends to an MKNF interpretation $M$ which induces $\hi$.
        \label{SAT:Completion:Saturation:Induces extends lemma}
        \item $\hi$ is induced by some MKNF interpretation $M$ such that $M \mmodels \kpo$.
        \label{prop:three}
        \item 
        $\hi \models \compsatr$.
        \item  
            $OB_{\ont, \hi} \nm a$
            for every atom $a \in (\KA \cup \{\bot\}) \setminus \hi$.
    \end{enumerate}
\end{lemma}

The relationship between saturated \ki s which satisfy the 
rule completion and MKNF interpretations is described by the following 
proposition.

\begin{proposition}
    \label{prop:rule and satur completion}
    For any \ki\ $\hi$ of an HMKNF-KB $\Kintro$,
    $\hi \models \pro_{rule} \wedge \ont_{satr}(\hi)$,
    if and only if,
    there exists an MKNF interpretation $M$
    such that $M$ induces $\hi$ 
    and $M \mmodels \pi(\K)$.
\end{proposition}

Unlike the rule completion, the support completion of \citep{loop_formulas}
is not immediately adaptable to HMKNF-KBs.
This is because their notion of support is
based on the idea that atoms must be implied by rules,
but applying this idea directly to HMKNF-KBS would be misguided
as they combine both rules based and ontological reasoning.
We instead define two different notions of support for an 
atom $a$ occurring within a \ki\ $\hi$ of an HMKNF-KB $\Kintro$.
Firstly, $a$ is supported by a rule $r \in \pro$,
if $\hi \models r$ while $\hi \setminus \{a\} \nm r$.
Secondly, $a$ is supported via the ontology $\ont$ if $OB_{\ont, \hi \setminus \{a\}} \models a$.

We can now specify a formula which ensures 
that all atoms within a satisfying \ki\ are supported 
by either a rule or the ontology.

\Needspace{8\baselineskip}
\begin{samepage}
\begin{definition}
    The {\em support completion} of an HMKNF-KB $\Kintro$, 
    w.r.t.\ a \ki\ $\hi$, 
    is the set of formulas
    $\compsup = \{\phi(a) \ | \ a \in \KA, OB_{\ont, \hi\setminus\{a\}} \nm a\}$,
    where
    \begin{align*}
        \phi(a) =  
            a \supset
            \bigvee_{\substack{
                r \in \pro \,\&\,
                a \in \rh
            }}
            \big(
                Body(r) \wedge 
                \bigwedge_{p \in \rh \setminus \{a\}}
                \neg p
            \big).
     \end{align*}
\end{definition}
\end{samepage}

When we combine the rule completion, saturation completion, 
and support completion, 
we obtain a formula which determines whether a \ki\ of 
a tight HMKNF-KB is induced by some MKNF model.

\begin{definition}
    The {\em completion} of an HMKNF-KB $\Kintro$
    with respect to a \ki\ $\hi$, 
    is $\kci = \comprule \wedge \compsatr \wedge \compsup$.
\end{definition}

\begin{theorem}
    \label{completion wrt I theorem}
    For any \ki\ $\hi$ of a tight HMKNF-KB $\Kintro$,
    $\hi \models \kci$,
    if and only if,
    $\K$ has an MKNF model $M$
    such that $M$ induces $\hi$.
\end{theorem}

\subsection{Loop Formulas}
\label{sec:loop formulas}
The loop formulas of \citep{loop_formulas} generalize the support completion and
allow for the assumption of tightness to be dropped by ensuring that each 
set of atoms forming a loop has support.
We generalize our earlier notion of support to any subset $L$,
of a \ki\ $\hi$ for an HMKNF-KB $\Kintro$ as follows.
A rule $r \in \pro$ supports $L$ if $\hi \models r$ 
but $\hi \setminus L \nm r$,
whereas $L$ is supported by the ontology $\ont$ 
if $OB_{\ont, \hi \setminus L} \models \bigvee L$.
Thus, the conversion of loop formulas to the case of HMKNF-KBs, 
is similar to that for the support completion, 
instead of requiring that each atom within a \ki\ is supported,
we require that each subset forming a loop within $\gk$ is supported
by either a rule or the ontology.

\begin{definition}
    The {\em loop formulas} of an HMKNF-KB $\Kintro$ 
    with respect to a \ki\ $\hi$,
    is the set of formulas
    $\kli = \{ \psi(L) \ | \ L \in \lk, OB_{\ont, \hi \setminus L} \nm \bigvee L \}$, where 
    \begin{align*}
        \psi(L) = 
            \bigvee L \supset 
            \bigg(
                    \bigvee_{\substack{r \in \pro \\ head(r)\cap L \ne \emptyset \\
                    body^+(r)\cap L = \emptyset}}
                        \Big(
                            Body(r)\wedge \bigwedge_{a\in head(r) \setminus L}
                            \neg a	
                        \Big) 
            \bigg).
    \end{align*} 
\end{definition}

We combine the completion and loop formulas
to determine whether a \ki\ of an arbitrary HMKNF-KB is induced by any MKNF model.

\needspace{2\baselineskip}
\begin{samepage}
\begin{theorem}
    \label{SAT:loop wrt I theorem}
    For any \ki\ $\hi$, of an HMKNF-KB $\Kintro$,
    $\hi \models \kci \wedge \kli$,
    if and only if,
    $\K$ has an MKNF model $M$ such that $M$ induces $\hi$. 
\end{theorem}
\end{samepage}
\begin{proof}[Proof Sketch]
    $(\Rightarrow)$
    It can be shown that whenever $\hi \models \comprule \wedge \compsatr$,
    it extends to an MKNF interpretation $M$ such that $M \mmodels \pk$.
    Note that $M$ also induces $\hi$ in this case.
    We must still show that there is no $M' \supset M$, 
    such that $\forall I' \in M', (I',M',M) \models \pk$.
    Therefore, we investigate the \ki\ $\hi'$, for which such an $M'$ induces.
    It can be shown that $\diffsetp \neq \emptyset$. 
    Thus we consider whether there is an atom $p \in \diffsetp$, 
    such that $p$'s truth is implied via a rule or the ontology under $\hi'$.
    We take $G$ to be the subgraph of $\gk$, 
    containing only atoms within $\diffsetp$
    which are reachable from an atom $g \in \diffsetp$.
    It is always possible to select an atom $g$, 
    such that either $G$ is acyclic 
    or $G$ contains only atoms from a single loop $L$.
    In the former case there is some atom 
    $p$, which has no outgoing edges in $G$. 
    The fact that $\hi \models \kci$ can then be shown to imply that 
    $p$ has a form of support within $\hi'$.
    In the later case, 
    all atoms in $L$ only have outgoing edges in $G$ to other atoms in $L$.
    The fact that $\hi \models \kli$ can be shown to imply that 
    some atom $p \in L$ has a from of support
    outside $L$ and within $\hi'$.
    In both cases this is sufficient to show 
    that $\forall I' \in M', (I',M',M) \nm \pk$, 
    and thereby show that $M$ is a model of $\K$.

    \noindent
    $(\Leftarrow)$
    This direction is relatively straightforward.
    It primarily relies on showing how 
    $M$ being a model of $\K$
    imposes restrictions on the \ki\ it induces.
\end{proof}

\begin{example}
    Consider the following HMKNF-KB $\Kintro$,
    representing whether a person $p$, 
    is a good candidate for a blood-pressure medication:
    \medskip
    \begin{align*}
        \pro =
        \left\{
            \begin{array}{ll}
            r_1 = goodCand(p) \leftarrow (cand(p), \neg highRisk(p)). & r_2 =highBP(p). \\
            r_3 = highRisk(p) \leftarrow (riskFactor(p), \neg risksTreated(p)). &  \\
            \end{array}
        \right\}
    \end{align*}
    \medskip
    and $\po = \{\forall x, (highBP(x) \supset cand(x)) \wedge (highRisk(x) \supset riskFactor(x))\}$.
    \medskip

    The ontology states
    that any person with high blood-pressure is a candidate, 
    and that anyone who is high-risk for the drug has a risk factor.
    The rule base states
    that $p$ is a good candidate for the drug if 
    they are a non high risk candidate, 
    that they are high risk if they have a risk factor 
    they have not been treated for, 
    and that they have high blood-pressure. 
    Below is the dependency graph $\gk$.

    
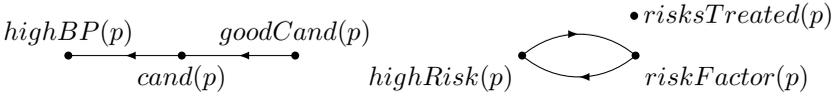
\bigskip
    \begin{tikzpicture}
        \tikzset{myarrow/.style={postaction={decorate},
        decoration={markings,
        mark=at position #1 with {\arrow{latex}},
        }}}
            
        \path
        (0,0) coordinate (a) node[above]{$highBP(p)$}--     
        +(0:1.5) coordinate (b) node[below]{$cand(p)$}--
        +(0:3) coordinate (c) node[above]{$goodCand(p)$}--
        +(0:6) coordinate (d) node[below left]{$highRisk(p)$}--
        +(0:7.5) coordinate (e) node[below right]{$riskFactor(p)$}
        +(4:7.5) coordinate (f) node[right]{$risksTreated(p)$}
        ;
        \draw[myarrow=.5] (b)--(a);
        \draw[myarrow=.5] (c)--(b);
        \draw[myarrow=.5] (d) to[bend left=40] (e);
        \draw[myarrow=.5] (e) to[bend left=40] (d);
        
        \foreach \p in {a,...,f}
        \fill (\p) circle(1.5pt);
    \end{tikzpicture}

    \bigskip 

    The following are selected \ki s for $\K$,
    of which only $\hi_4$ is induced by any MKNF model: 
    $\hi_1 = \{highBP(p)\}$,
    $\hi_2 = \{highBP(p),$ $cand(p), goodCand(p),$ $risksTreated(p)\}$,
    $\hi_3 = \{highBP(p),$ $cand(p),$ $highRisk(p),$ $riskFactor(p)\}$,
    and $\hi_4 = \{goodCand(p),$ $cand(p),$ $highBP(p)\}$. 

    Clearly, $cand(p)$ is entailed by the ontology given $\hi_1$, 
    so $\hi_1$ fails to satisfy $\ont_{satr}(\hi_1)$.
    For $\hi_2$, it is clear that $risksTreated(p)$
    has no form of support, 
    therefore it fails to satisfy $\K_{sup}(\hi_2)$. 
    Finally, from examination of $\gk$, 
    $\{highRisk(p), riskFactor(p)\}$ 
    forms a loop $L \in \lk$.
    As each atom is only supported by the other
    $\hi_3$ does not satisfy $\K_{loop}(\hi_3)$.
    
    \comment{
        Since $OB_{\ont, \hi_{1}} \models cand(p)$,
        $cand(p) \in \ont_{satr}(\hi_1)$, 
        as $\hi_1 \nm cand(p)$,
        $\hi_1 \nm \K_{comp}(\hi_1)$.
        Clearly, $OB_{\ont, \hi_2 \setminus \{risksTreated(p)\}} \nm risksTreated(p)$,
        therefore there is a clause in $\K_{sup}(\hi_2)$ which 
        implies that since $\hi_2 \models risksTreated(p)$  
        there is some rule $r \in \pro$ such that 
        $\hi_2 \models Body(r)$ and $risksTreated(p)$ is 
        the only atom $a \in \rh$ where $\hi_2 \models a$. 
        There is no such rule so $\hi_{2} \nm \K_{comp}(\hi_2)$.
        From examination of $\gk$, $L = \{highRisk(p), riskFactor(p)\} \in \lk$.
        Consequently, as $OB_{\ont, \hi_3 \setminus L} \nm \bigvee L$, 
        there is a clause in $\K_{loop}(\hi_3)$ which implies that since 
        $\hi_3 \models \bigvee L$ there is 
        some rule $r \in \pro$ where
        $\prb \cap L = \emptyset$ and 
        $\rh \cap L \neq \emptyset$ 
        such that  
        $\hi_3 \models Body(r)$, and 
        for any atom $a \in \rh$ where 
        $\hi_3 \models a$, $a \in L$.
        For the only rule $r\in \pro$ such that
        either atom from $L$ is in $\rh$,
        $riskFactor(p) \in \prb$,
        therefore $\hi_3 \nm \K_{loop}(\hi_3)$. 
    }

    In contrast to the other \ki s,
    $\hi_4$ satisfies $\K_{comp}(\hi_4)$ and $\K_{loop}(\hi_4)$. 
    It avoids the problem with $\hi_1$ as $\hi_4 \models cand(p)$, 
    the one with $\hi_2$ as $\hi_4 \nm risksTreated(p)$, and 
    the one with $\hi_3$ as $\hi_4 \nm \bigvee L$.
    This is consistent with the fact 
    that it is the only \ki\ induced by an MKNF model.

\end{example}

\section{Nogoods}
\label{sec:nogoods}
In what follows,
we present sets of nogoods
indirectly capturing the constraints induced by the completion and loop formulas
of the previous section.
Total assignments of these nogoods directly correspond with \ki s, 
and their solutions to those induced by MKNF models.
True atoms within an assignment reflect those 
evaluated as true under the corresponding \ki, 
and similarly for false atoms.
Through this relationship the nogoods characterize MKNF models.
As such, 
conflict-driven approaches can be built on generating a 
subset of these nogoods.

In all definitions of this section, we assume a given HMKNF-KB $\Kintro$.

\subsection{Completion Nogoods}
\label{sec:Completion Nogoods}

\noindent
{\bf Rule Nogoods:}

For expressing that the body of a rule $r$ is satisfied, \cite{gebser2013advanced}
use sets of literals of the form $\beta(r) = \{\T p \ | \ p \in \prb\} \cup \{\F p \ | \ p \in \nrb\}$.
These are treated as composite variables with an intrinsic meaning.
The literal $\T \beta(r)$ represents $Body(r)$ being satisfied
whereas $\F \beta(r)$ represents its unsatisfaction.
This meaning is enforced within solutions 
by a set of \emph{conjunction nogoods} to be introduced shortly.

Directly following \citep{gebser2013advanced}, 
we define a set of \emph{rule nogoods} corresponding to the rule completion, 
which in our case ensures that the rule base is satisfied.

\begin{definition}
    The {\em rule nogood} for any $r \in \pro$, is defined as: $\phi_{\pro}(r)=\{\F p_1,\ldots, \F p_t, \T \beta(r) \ | \ head(r)= \{p_1, \ldots, p_t \} \}.$
    The {\em rule nogoods} of $\K$ are $\Phi_{\pro} = \{\phi_{\pro}(r) \ | \ r \in \pro \}$.
\end{definition}

\smallskip
\noindent
{\bf Saturation Nogoods:}

To represent whether an atom $p$ is 
supported by the ontology, 
we use the variable $\beta_\ont(p)$.
Within an assignment,
$\T \beta_{\ont} (p)$ represents that
$p$ is supported via $\ont$,
and $\F \beta_{\ont} (p)$ represents that it is not.
This is enforced within solutions 
by a set of \emph{entailment nogoods} to be introduced shortly.

To express that a solution must be reflective of a saturated \ki, 
we introduce a novel set of \emph{saturation nogoods} 
corresponding to the saturation completion. 

\begin{definition}
    The {\em saturation nogood} for any atom $p \in \KAO \cup \{\bot\}$,
    is defined as: 
    $\phi_{\ont}(p) = \{\F p, \T \beta_{\ont} (p)\}$.
    The {\em saturation nogoods} of $\K$ are $\Phi_{\ont} = \{\phi_{\ont}(p) \ | \ p \in \KAO \cup \{\bot\}\}$.
\end{definition}

\medskip
\noindent
{\bf Support Nogoods:}

To reflect whether an atom $p$
is supported by a rule $r$, 
we use literal sets
of the form
$\beta_{\pro}(r,p) = \{\T \beta (r)\} \cup \{\F q \ | \ q \in \rh \setminus \{p\}\}$,
as is done in \citep{gebser2013advanced}.\footnote{
    In \citep{gebser2013advanced} 
    literal sets denoted as $\beta_{\pro}(r,p)$, 
    are defined such that they are substituted for 
    by other literal sets of the form $\beta(r)$, 
    whenever they effectively coincide. 
    This is a relevant consideration for implementation efficiency; 
    however it is omitted here for simplicity.}
The truth of literals based on these sets are 
also enforced by conjunction nogoods.

We introduce a novel set of \emph{support nogoods} 
corresponding to the support completion,
to prevent cases where there 
is no support from a rule or the ontology 
for some true atom within an assignment.

\begin{definition}
    The {\em support nogood} for any atom $p \in \KA$, is defined as
    \begin{align*}
        \psi_{\K} (p) = \{\T p\} \cup \{\F \beta_{\pro} (r,p) \ | \ r \in \pro, p \in \rh\}
        \cup \{\F \beta_{\ont} (p) ~|~ \textrm{if } p \in \KAO\}.
    \end{align*}
    \comment{
    and for any atom $p \in (\KA \setminus \KAO)$, is defined as
    \begin{align*}
        \psi_{\K} (p) = \{\T p\} \cup \{\F \beta_{\pro} (r,p) \ | \ r \in \pro, p \in \rh\}.
    \end{align*} 
    }
    The {\em support nogoods }of $\K$, are $\Psi_{\K} = \{\psi_{\K}(p) \ | \ p \in \KA\}$.
\end{definition}

\smallskip
\noindent
{\bf Conjunction Nogoods:}

As aforementioned, conjunction nogoods are required to ensure 
that composite variables consisting of 
other literals are assigned correctly within solutions. 
Here, we directly follow \citep{gebser2013advanced}.

\begin{definition}
    The {\em conjunction nogoods} for a set of literals $\beta$ are 
  $ \gamma_{\pro}(\beta) = \{\{\F \beta\} \cup \beta \} \cup \{\{\T \beta, \overline{\sigma}\} \ | \ \sigma \in \beta\}$.
    The {\em conjunction nogoods} of $\K$ are 
    $\Gamma_{\pro} = 
    \bigcup_{\beta \in \{\beta(r) | r \in \pro\} \cup \{\beta_{\pro}(r,p) | r\in \pro, p \in \rh\}}
    \gamma_{\pro}(\beta).$
\end{definition}


\smallskip
\noindent
{\bf Entailment Nogoods:}

To ensure the literals of the form 
$\T \beta_{\ont} (p)$ or $\F \beta_{\ont} (p)$ 
appear within assignments in accordance with 
whether $p$ is supported via the ontology,
we introduce \emph{entailment nogoods}.

\begin{definition}
  
  Let $\beta_{\ont}(p)$ be a variable 
    associated with an atom $p \in \KAO$.
    A {\em positive entailment nogood}, 
    of an atom $p\in\KAO$ and set $S \subseteq \KAO$,
    is defined as
    $\gamma_{\ont}^+ (p, S) = \{\F \beta_{\ont}(p)\} \cup \{\T s \ | \ s \in S\}$,
    whereas a {\em negative entailment nogood}, is defined as 
    $\gamma_{\ont}^- (p, S) = \{\T \beta_{\ont}(p)\} \cup \{\F s \ | \ s \in S\}.$
    The {\em entailment nogoods }of $\K$ are 
    \comment{
        \begin{align*}
            \Gamma_{\ont} = 
            \bigcup_
            {\substack{
                p \in \KAO \cup \{\bot\},S \subseteq \KAO, \\ OB_{\ont, S \setminus \{p\}} \models p
                }
            }
            \{\gamma_{\ont}^+ (p, S)\}
            &\cup
            \bigcup_
            {\substack{
                p \in \KAO,S \subseteq \KAO, \\ OB_{\ont, \KAO \setminus (S \cup \{p\})} \nm p
                }
            } 
            \{\gamma_{\ont}^- (p, S)\}.
        \end{align*}
    }
    \begin{align*}
        \Gamma_{\ont} = &
        \{\gamma_{\ont}^+ (p, S) \ | \ p \in \KAO \cup \{\bot\},S \subseteq \KAO, OB_{\ont, S \setminus \{p\}} \models p\} \\
        &\cup
        \{\gamma_{\ont}^- (p, S) \ | \ p \in \KAO,S \subseteq \KAO, OB_{\ont, \KAO \setminus (S \cup \{p\})} \nm p \}.
    \end{align*}
    For some atom $p \in \KAO$ and set of atoms $S \subseteq \KAO$,
    the purpose of a positive entailment nogood $\gamma_{\ont}^+(p,S)$
    is to indicate that $p$ is supported via $\ont$
    by true atoms within an assignment,
    given it has all atoms in $S$ as true.
    Conversely, 
    the purpose of a negative entailment nogood
    $\gamma_{\ont}^-(p,S)$
    is to indicate that $p$ 
    has no way of being supported via $\ont$
    by true atoms within an assignment,
    given it has all atoms in $S$ as false.

    \medskip
    \noindent
    {\bf Relation to MKNF Models:}

    So far we have discussed the relationship between assignments 
    and \ki s using general language. 
    The exact correspondance is given by the following definition.

    \begin{definition}
        Let $\hi$ a \ki\ for $\K$. The induced assignment of $\hi$ for $\K$ is
        \begin{align*}
            &A_{\K}^{\hi} = \{\T p \ | \ p \in \hi\} \cup
            \{\F p \ | \ p \in \KA \setminus \hi\} 
            \cup \{\F \bot\}
            \nonumber \\ 
            &\cup \{\T \beta(r) \ | \ r \in \pro, \hi \models Body(r)\} 
            \cup \{\F \beta(r) \ | \ r \in \pro, \hi \nm Body(r)\} 
            \nonumber \\
            &\cup \{\T \beta_{\pro} (r,p) \ | \ r \in \pro, \hi \models Body(r), \rh \cap (\hi \cup \{p\}) = \{p\}\}
            \nonumber \\
            &\cup \{\F \beta_{\pro} (r,p) \ | \ r \in \pro, \hi \nm Body(r), p \in \rh\}
            \nonumber \\
            &\cup \{\F \beta_{\pro} (r,p) \ | \ r \in \pro, p \in \rh, \rh \cap \hi \not \subseteq \{p\} \} 
            \nonumber \\
            &\cup \{\T \beta_{\ont} (p) \ | \ p \in \KAO \cup \{\bot\}, OB_{\ont, \hi \setminus \{p\}} \models p \}
            \nonumber \\
            &\cup \{\F \beta_{\ont} (p) \ | \ p \in \KAO \cup \{\bot\}, OB_{\ont, \hi \setminus \{p\}} \nm p\}.
        \end{align*}
    \end{definition}

    The nogoods we have defined thus far make up our \emph{completion nogoods}.
    \begin{definition}
        The {\em completion nogoods} of $\pk$ are
        \begin{align*}
            \Delta_{\K} = 
            \Phi_{\pro} \cup \Phi_{\ont} \cup \Psi_{\K} \cup 
            \Gamma_{\pro} \cup \Gamma_{\ont}.
        \end{align*}
    \end{definition}
    We have designed the induced assignment of any \ki\ 
    to be a total assignment for the completion nogoods,
    and the nogoods themselves to occur in and only in
    assignments induced by \ki s which do not satisfy their completion formulas.
    In showing this to be the case, we obtain the following result.

    \begin{theorem}
        \label{nogood comp theorem}
        Let $\hi$ be a \ki\ for an HMKNF-KB $\Kintro$,
        then we have that
        $\hi \models \kci$ if and only if $\AI$ is
        a solution to $\Delta_{\K} \cup \{\T \bot\}$.
    \end{theorem}

    The above theorem allows us to conclude that
    the \ki s that induce solutions to the completion nogoods
    are the ones which sastisfy their completion formulas. 
    It naturally follows from \Theorem{completion wrt I theorem},
    that for tight HMKNF-KBs
    these are also the \ki s induced by MKNF models.

    \comment{
    \begin{corollary}
        Let $\hi$ be a \ki\ for a tight HMKNF-KB $\Kintro$,
        then we have that $\hi$ is induced by an MKNF model of $\pk$ iff $\AI$
        is a solution to $\Delta_{\K} \cup \{\T \bot\}$.
    \end{corollary}
    }

\end{definition}

\subsection{Loop Nogoods}
    \label{sec:loop nogoods}
    
    \emph{Loop nogoods} are intended to parallel the 
    loop formulas of \Section{sec:loop formulas}
    in ensuring non-circular support.
    Directly following \citep{gebser2013advanced},
    we denote the external program supports of a
    set of atoms $L$,
    by $\epsi_{\pro} (L) = \{r \in \pro \ | \
    \rh \cap L = \emptyset, \prb \cap L \neq \emptyset\}$. 
    Moreover, $\rho(r, L) =
    \{\F \beta(r)\}\cup \{\T p \ | \ p \in \rh \setminus L\}$,
    collects all literals satisfying a rule $r$,
    regardless of whether any atom from $L$ is true.

    \begin{definition}
        The {\em loop nogoods} for any $L \subseteq \KA$ and $S \subseteq \KAO$ are defined as
        \begin{align*}
            \lambda_{\K} (L, S) =  \{&
                \{\T p, 
                \sigma_1,\ldots, \sigma_k,
                \F s_1, \ldots, \F s_n \}
                \ | \
                p \in L,
                \epsi_{\pro}(L)= \{r_1,\ldots,r_k\}, \\
                &\sigma_1 \in \rho(r_1,L),\ldots,\sigma_k \in \rho(r_k, L), 
                S = \{s_1, \ldots, s_n\}
                \}.
        \end{align*}
        The loop nogoods of $\K$ are 
        \begin{align*}
                \Lambda_{\K} =
                \bigcup_{\substack{
                        L \in \lk, S \subseteq \KAO, \\
                        L\cap S = \emptyset, OB_{\ont, \KAO \setminus (S \cup L) \nm \bigvee L}
                    }}
                    \lambda_{\K} (L, S).
        \end{align*}
    \end{definition}
    
    Above, $\Lambda_{\K}$ includes a nogood for 
    each loop, $L$, and subset of $\KAO$, $S$, 
    for which $L$ cannot be supported
    via $\ont$ whenever all atoms in $S$ are false.
    Each nogood $\lambda(L,S)$ requires that the loop $L$ 
    must be supported by a rule if all atoms in $S$ are false.
    We can show that loop nogoods occur only 
    within assignments induced by 
    \ki s which do not satisfy their loop formulas,
    to obtain the following result.
    \begin{theorem}
        \label{nogood loop theorem}
        Let $\hi$ be a \ki\ for an HMKNF-KB $\Kintro$,
        then we have that
        $\hi \models \kci \wedge \kli$ if and only if
        $\AI$ is a solution to $\dk \cup \lnk \cup \{\T \bot\}$.
    \end{theorem}
    \Needspace{8\baselineskip}
    \begin{samepage}
    \begin{proof}[Proof Sketch]
        $(\Rightarrow)$ It can be shown relatively easily 
        that $\AI$, the assignment $\hi$ induces,
        is a total assignment for the nogoods $\dk \cup \lnk \cup \{\bot\}$.

        The other condition which must be shown 
        is that no nogood from $\dk \cup \lnk \cup \{\bot\}$
        occurs within $\AI$.
        This is done by considering each of
        $\Phi_{\pro}$, $\Phi_{\ont}$, $\Psi_{\K}$, 
        $\Gamma_{\ont}$, $\Gamma_{\pro}$, and $\lnk$ 
        individually.
        For each of nogood set $\nabla$ we make the assumption that 
        $\delta \in \nabla$ is a subset of 
        $\AI$, then show that the restrictions this imposes on $\hi$
        also imply that for some literal $\sigma \in \delta$,
        $\overline{\sigma} \in \AI$.
        This contradicts that $\AI$ is a total assignment 
        and in doing so proves no such nogood can exist.

        For example:
        Assume there is a nogood $\gop(p,S) \in \Gamma_{\ont}$ 
        for $p \in \KAO \cup \{\bot\}$,
        $S \subseteq \KAO$ such that $\gop(p,S) \subseteq \AI$.
        Then by the fact that $\gop(p,S) \in \Gamma_{\ont}$,
        $OB_{\ont, S \setminus \{p\}} \models p$, 
        moreover since $\gamma_{\ont}^+ (p, S) = \{\F \beta_{\ont}(p)\} \cup \{\T s \ | \ s \in S\}$ 
        and $\gop(p,S) \subseteq \AI$, $S \subseteq {\AI}$.
        Therefore $OB_{\ont, \hi \setminus \{p\}} \models p$ as well,
        so $\T \beta_{\ont}(p) \in \AI$.
        This contradicts the fact that $\AI$ is a total 
        assignment proving that no such $\gop(p,S)$ exists.
        The inverse logic can be applied to show that no nogood
        $\gon(p,S) \in \Gamma_{\ont}$ for $p \in \KAO$,
        $S \subseteq \KAO$ can occur in $\AI$.

        \comment{
            Another example:
            Based on the construction of $\AI$ for any rule $r \in \pro$,
            $\forall p \in \prb, \T p \in \AI$ and $\forall n \in \nrb, \F p \in \AI$,
            if and only if $\hi \models Body(r)$.
            Thus if $\T \beta (r) \in \AI$ if each literal within $\beta(r)$ 
            is within $\AI$ and otherwise $\F \beta(r) \in \AI$.
            Consequently, no nogood contained within a 
            set of the form $\gamma_{\pro}(\beta(r))\in \Gamma_{\pro}$ 
            may occur within $\AI$.
            If we instead assume that some nogood $\delta$ within a set of the form
            $\gamma_{\pro}(\beta(r, p)) \in \Gamma_{\pro}$
            such that $\delta \subseteq \AI$ we also run into a contradiction.
            
            Assume some that there is a nogood $\delta \in \Gamma_{\pro}$,
            such that $\delta \subseteq \AI$.
            It can be shown that for this to be the case there must be some 
            variable $\sigma$ for which $\T \sigma$ and $\F \sigma$ are in $\AI$
            which contradicts the fact that $\AI$ is a total assignment.
            For example if $\{\T \beta(r), \F p\} \subseteq \AI$ for some rule 
            $r$ where $p \in \prb$,
            then because $\T \beta(r) \in \AI$, $I \models Body(r)$,
            so $I \models p$, 
            and therefore $\T p \in \AI$.}
            \medskip
        \noindent
        $(\Leftarrow)$ 
        We wish to show that $\hi$ must satisfy 
        all of $\comprule$, $\compsatr$, $\compsup$, and
        $\kli$.
        To do so we consider the case where $\hi$ fails 
        to satisfy each of them individually, 
        and show that it would require a nogood 
        to exist within $\AI$,
        violating the inital assumptions. 

        For example:
        Assume that $\hi \nm \compsup$.
        Clearly there is some atom $p \in \hi$ 
        such that $OB_{\ont, \hi \setminus \{p\}} \nm p$
        for which for all $r \in \pro$ where $p \in \rh$ 
        $\hi \nm Body(r)$ or $\exists a \in \rh\setminus\{p\}$ 
        such that $a \in \hi$. 
        From $p \in \hi$, $\T p \in \AI$.
        From $OB_{\ont, \hi \setminus \{p\}} \nm p$,
        either $p \in (\KA \setminus \KAO)$
        or $\F \beta_{\ont} \in \AI$.
        Finally, since for all $r \in \pro$ where $p \in \rh$,
        $\hi \nm Body(r)$ or $\exists a \in \rh\setminus\{p\}$ 
        such that $a \in \hi$, 
        $\F \beta_{\pro} (r,p) \in \AI$.
        It follows that $\psi_{\K} (p) \subseteq \AI$,
        since $\psi_{\K} (p) \in \Psi_{\K}$ 
        this violates the assumption that $\AI$ is a solution for $\Delta_{\K}$.
        Therefore $\hi \models \compsup$.
    \end{proof}
    \end{samepage}

    This theorem, together with \Theorem{SAT:loop wrt I theorem}, 
    implies that, for any HMKNF-KB,
    the K-interpretations that induce solutions
    to the completion and loop nogoods are precisely 
    those that are induced by MKNF models.
    \bigskip
    \begin{example}
        Consider the following HMKNF-KB $\Kintro$ where only the 
        \ki\ $\hi = \{a, b\}$ is induced by any MKNF model.
        \begin{align*}
            \pro =
            \left\{
                \begin{array}{ll}
                r_1 = a. & r_2 = a \vee d. \\
                r_3 = f \leftarrow d. & r_4 = e \leftarrow f.
                \end{array}
            \right\}
        \end{align*}
        and 
        $\po = \{(a \supset b) \wedge (c \supset d) 
                    \wedge (c \supset e) \wedge (e \supset f)\}$.
        $\gk$ is shown below.
\medskip

        
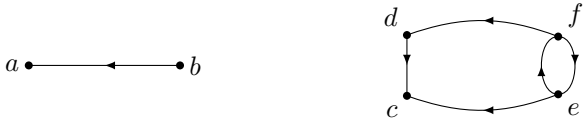
\begin{tikzpicture}
            \tikzset{myarrow/.style={postaction={decorate},
            decoration={markings,
            mark=at position #1 with {\arrow{latex}},
            }}}
                
            \path
            (0,0) coordinate (x)
            (2.5,0) coordinate (a) node[left]{$a$}--     
            (4.5,0) coordinate (b) node[right]{$b$}--
            (7.5,-0.40) coordinate (c) node[below left]{$c$}--
            (7.5,0.40) coordinate (d) node[above left]{$d$}--
            (9.5,-0.38) coordinate (e) node[below right]{$e$}--
            (9.5,0.38) coordinate (f) node[above right]{$f$}
            ;
            \draw[myarrow=.5] (b)--(a);
            \draw[myarrow=.5] (f) to[bend left=90] (e);
            \draw[myarrow=.5] (e) to[bend left=90] (f);
            \draw[myarrow=.5] (f) to[bend right=20] (d);
            \draw[myarrow=.5] (e) to[bend left=20] (c);
            \draw[myarrow=.5] (d)--(c);

            \foreach \p in {a,...,f}
            \fill (\p) circle(1.5pt);
        \end{tikzpicture}
       
        We will show that $\AI$, the assignment $\hi$ induces, 
        is the only solution to our nogoods.
        To do so we take $A$ to be an arbitrary solution and prove that it 
        contains every literal within $\AI$.
        Some steps are omitted for conciseness,
        for instance we leave it as an exercise to the reader to show that:
        \begin{align*}
            \AI = \{&\T a, \T b, \F c, \F d, \F e, \F f, \\
                    &\T \beta(r_1) = \T \beta(r_2) = \T \emptyset, 
                    \F \beta(r_3) = \F \{d\}, \F \beta(r_4) = \F \{f\}, \\
                    &\T \beta(r_1, a) = \T \{ \T \beta(r_1)\},
                    \T \beta(r_2, a) = \F \{ \T \beta(r_2), \F d\},  \\
                    &\F \beta(r_2, d) = \F \{ \T \beta(r_2), \F a\}, 
                    \F \beta(r_3, f) = \F \{ \T \beta(r_3)\}, 
                    \F \beta(r_4, f), = \F \{ \T \beta(r_4)\},\\
                    &\F \beta_{\ont} (a), \T \beta_{\ont} (b),
                    \F \beta_{\ont} (c), \F \beta_{\ont} (d),
                    \F \beta_{\ont} (e), \F \beta_{\ont} (f) \}
        \end{align*}
    \end{example}

    Firstly, we show that $\T a$ and $\T b$ must be in $A$.
    Consider the conjunction nogood 
    $\gamma(r_1) = 
    \{\{\F \beta(r_1)\} \cup \beta(r_1) \} \cup 
    \{\{\T \beta(r_1), \overline{\sigma}\} 
    \ | \ \sigma \in \beta(r_1)\} 
    \in \Gamma_{\pro}$.
    As the body of $r_1$ is empty $\beta(r_1) = \emptyset$, 
    thus $\gamma(r_1) = \{\F \emptyset\}$,
    and so $\T \emptyset \in A$.
    Comined with the rule nogood $\phi_{\pro}(r_1) = \{ \F a, \T \beta(r_1)\} \in \Phi_{\pro}$, 
    this also means that $\T a \in A$.
    Due to the fact that $OB_{\ont, a} \models b$, 
    the positive entailment nogood $\gop(b, \{a\})$
    occurs within $\Gamma_{\ont}$. As $\gop(b, \{a\}) = \{\F \beta_{\ont} (b), \T a\}$
    it follows that $\T \beta_{\ont}(b) \in A$.
    Consequently, due to the saturation nogood 
    $\phi_{\ont}(b) = \{\F b, \T \beta_{\ont}(b)\} \in \Phi_{\ont}$,
    $\T b \in A$ as well.

    It is simple to show that $\F c \in A$, 
    as such we focus on the more interesting case of $\F d$.
    Turn your attention to the conjunction nogoods 
    $\gamma_{\pro} (\beta_{\pro} (d, r_2)) \subseteq \Gamma_{\pro}$,
    which are $\{\F \beta_{\pro} (d, r_2), \T \beta(r_2), \F a\}$,
    $\{\T \beta_{\pro}(d,r_2), \F \beta(r_2)\}$,
    and $\{\T \beta_{\pro}(d,r_2), \T a\}$.
    The last one shows that $\F \beta_{\pro}(d,r_2) \in A$.
    Also note that unless $c$ is true $d$ cannot be 
    supported by the ontology 
    \-- $OB_{\ont, \KAO \setminus \{c, d\}} \nm d$ \--
    therefore the negative entailment nogood $\gon (d, \{c\}) = \{\T \beta_{\ont} (d), \F c\}$ 
    is in $\Gamma_{\K}$ and thereby $\F \beta_{\ont} (d) \in A$.
    Now consider the support nogood 
    $\psi_{\K}(d) = 
    \{\T d, \F \beta_{\pro}(d, r_2), \F \beta_{\ont}(d)\} \in \Psi_{\K}$,
    to see that $\F d \in A$.

    Finally we show that $\F e$ and $\F f$ are in $A$.
    Clearly the set of both atoms is a loop $\{e, f\} \in \lk$,
    and since neither atom can be supported by the ontology unless 
    $c$, $e$, or $f$ is true 
    \-- $OB_{\ont, \KAO \setminus \{c, e, f\}} \nm \bigvee \{e, f\}$ \--
    the loop nogoods $\lambda(\{e, f\}, \{c\})$ are in $\Lambda_\K$. 
    The external supports for $\{e, f\}$ are simply 
    $\epsi(\{e,f\}) = \{r_3\}$,
    and $\rho(r_3, \{e, f\}) = \{\F \beta(r_3)\}$,
    therefore 
    $\lambda(\{e, f\}, \{c\}) = 
    \{\{\T e, \F \beta(r_3)\}, \{\T f, \F \beta(r_3)\}\}$.
    From the fact that $\F d \in A$, 
    it is easy to show that $\F \{d\} = \F \beta(r_3) \in A$.
    Consequently $\F e$ and $\F f$ both must be in $A$.

    We have that $\{\T a, \T b, \F c, \F d, \F e, \F f\} \subseteq A$
    which is the most essential aspect of $\AI$'s relationship to $\hi$,
    the rest is left as an exercise.

\section{Conflict-Driven Solving}
A conflict-driven solver
which determines \ki s induced by MKNF models of an HMKNF-KB,
can be built based on the completion and loop nogoods of \Section{sec:nogoods},
following the same general approach of \citep{gebser2012}.
The following is a sketch of such a solver.
Our goal is to provide an overview 
which is open to further specification.

\label{sec:conflict}
\subsection{Main Procedures}
Due to the similarity of their formulation to that of \citep{gebser2012}
the details of the main conflict-driven procedures
will be explained only at a high level,
with some comments on specific differences.

{\bf Algorithm \ref{algo 1}:}\emph{ CDNL} is primarily 
responsible for keeping track of an 
assignment representing a partial candidate solution,
through a tree-like search procedure. 
Additionally, it tracks a set of nogoods
which are a subset of those introduced in \Section{sec:nogoods}, 
and a decision level indicating the 
current depth of the search tree.
It operates by calling the reasoning procedure \emph{NogoodProp}
to deterministically extend the assignment,
and track additional nogoods 
which prove helpful in doing so.
Following this it takes one of three actions.
If the current assignment is incomplete
but compatible with the current nogood,
an arbitrary decision literal is added to the assignment
increasing the decision level.
If the current assignment conflicts 
with the current nogoods,
the search backtracks to a lower decision level
or returns `no model' if already at the lowest decision level.
Finally, if the assignment is total and compatible with all tracked nogoods, 
a model-checking oracle is consulted. 
If the assignment is verified as a solution the 
\ki\ that induces it is returned. 
Otherwise, the trivial nogood
\-- the full invalid assignment \--
is added to the tracked set enabling backtracking.

The significant changes from the main algorithm of \citep{gebser2012} are:
the different initial set of nogoods,
and the final model check
ultimately required for the soundness of the algorithm.
This check can only be avoided
if we assume we can always determine a 
nogood whenever one is a subset of a total assignment,
however this requires substantial
procedures for determining loop nogoods.

{\bf Algorithm \ref{algo 2}: }\emph{NogoodProp} 
is the core reasoning procedure.
A unit-resulting nogood for an assignment contains 
one literal which is not within the assignment.
We use the term unit-propagation to refer to the 
process of adding the complement of this literal 
to the assignment which the nogood is unit-resulting for.
\emph{NogoodProp} unit-propagates the current assignment
based on the set of tracked nogoods, 
and adds additional nogoods whenever propagation reaches a fixpoint. 
The novel feature of this procedure
is that its first resort
after reaching a fixpoint
is to call \emph{EntNogoods},
which will attempt to add unit-resulting 
entailment nogoods to $\nabla$. 
If no such nogoods can be found based on the current assignment, 
the procedure will instead attempt to generate a loop nogood, 
by calling the procedure \emph{UnfoundedSet}. 
This procedure returns a set $U \in \lk$\footnote{
    The condition that $U$ be in $\lk$ is relaxable.
    As the dependency graph $\gk$
    can contain extra edges 
    without affecting the completeness
    of \Theorem{SAT:loop wrt I theorem},
    any $S \subseteq \KA$ may be added to $\lk$.
} 
for which $U \subseteq A^T$,
and $S \subseteq A^F$ such that $OB_{\ont, \KA \setminus (U \cup S)} \nm \bigvee U$.
In principle, this can be a modification of existing 
methods for detecting unfounded sets.
Eventually, it reaches a point where either
a tracked nogood conflicts with the current assignment
or no more unit-resulting nogoods can be discovered.
At this point, the procedure returns.

\begin{figure}[htbp]
\centering

\begin{minipage}[t]{.495\textwidth}
\footnotesize
\begin{algorithm}[H]
\DontPrintSemicolon
\SetAlgoVlined
\label{algo 1}

let $\Delta$ be $\Phi_{\pro} \cup \Phi_{\ont} \cup \Psi_{\K} \cup \{\bot\}$

$(\nabla, A, dl) \leftarrow (\emptyset, \{\F \bot\}, 0)$

\BlankLine

\While{True}{
    $(A, \nabla) \leftarrow $ \textit{NogoodProp}$(dl,\K,\nabla,A)$

    \If{$\epsi \subseteq A$ for some $\epsi \in \Delta \cup \nabla$}
    {
        \If{$dl = 0$}
        {
            \Return{no model}
        }

        $(\delta, dl) \leftarrow$ \textit{ConfAnal}$(\epsi, \K, \nabla, A)$
        $\nabla \leftarrow \nabla \cup \{\delta\}$
        $A \leftarrow A \setminus \{\sigma \in A \ | \ dl < dl(\sigma)\}$ 
    }
    \ElseIf{$A^T \cup A^F$ is total}
    {  
        \If{ModelCheck(A)}
        {
            \Return{$A^T \cap \KA$}
        }
        \Else{
            $\nabla \leftarrow \nabla \cup A$
        }
    }
    \Else
    {
        $\sigma_d \leftarrow Select(\K, \nabla, A)$

        $dl \leftarrow dl + 1$

        $dl(\sigma_{d}) \leftarrow dl$

        $A \leftarrow A \circ \sigma_d$
    }
}
\caption{\it{CDNL}}
\end{algorithm}
\end{minipage}\hfill
\begin{minipage}[t]{.495\textwidth}
\footnotesize
\begin{algorithm}[H]
\label{algo 2}
\DontPrintSemicolon
\SetAlgoVlined

$U \leftarrow \emptyset$

\BlankLine

\While{True}{
    \If{$\delta \subseteq A$ for some $\delta \in \Delta \cup \nabla$}
    {
        \Return{$(A,\nabla)$}
    }

    $\Sigma \leftarrow \{\delta \in \Delta \cup \nabla \ | \ \delta \setminus A = \{\overline{\sigma}\}, \sigma \not \in A \}$

    \If{$\Sigma \neq \emptyset$}
    {
        let $\overline{\sigma} \in \delta \setminus A$ for some $\delta \in \Sigma$

        $dl(\sigma) \leftarrow dl$

        $A \leftarrow A \circ \sigma$
    }
    $(\nabla, $\textit{ent}$) \leftarrow EntNogoods(A, \K, \nabla)$ 
    
    \If{\textbf{\emph{not}} ent}
    {
        $U \leftarrow U \setminus A^F$

        \If{$U = \emptyset$}
        {
            $(U, S) \leftarrow UnfoundedSet(\K, A)$
        }

        \If{$U = \emptyset$}
        {
            \Return{$(A, \nabla)$}
        }

        \Else
        {
            let $p\in U, \delta \in \lambda_{\K}(p,U,S)$

            $\nabla \leftarrow \nabla \cup \delta$
        }
    }
}
\caption{\it{NogoodProp}}
\end{algorithm}
\end{minipage}
\end{figure}
\subsection{Determining Entailment Nogoods}
{\bf Algorithm \ref{algo 3}: }\emph{EntNogoods}
is the novel procedure responsible for 
determining unit-resulting entailment nogoods.
It relies on the function \textit{Entailment},
which takes a set $S \subseteq \KAO$ as an argument.
It returns the set $\{\bot\}$ if $OB_{\ont, S}$ is inconsistent,
and otherwise returns a set $\Omega = \{p \in \KAO  \ | \ OB_{\ont, S} \models p\} \cup \{\neg p \ | \ p \in \KAO, OB_{\ont, S} \models \neg p\}$. 
We let $\Omega^+$ refer to $\{p \in \Omega \ | \ p \in \KAO \cup \{\bot\}\}$
and $\Omega^-$ refer to $\{\neg p \in \Omega \ | \ p \in \KAO\}$.
We denote the set of atoms which some $p \in \KAO$
has an edge to in $\go$ as $ext(p)$.

The procedure determines the minimum set of entailable information $\Omega$ 
in \Line{algo:lower bound},
by calling \emph{Entailment} 
with all atoms from $\KAO$ which are true in the current assignment.
It then checks if this set contains the contradiction atom $\bot$
in \Line{algo:if bot}.
If so it adds the appropriate nogood, 
and returns indicating success.
Otherwise, it checks whether any atom
not currently known to be true
can be entailed as such in \Line{algo:if true}.
If so it adds a nogood for each such atom, 
allowing the atom to be unit-propogated.
Then in \Line{algo:add true},
the generated nogood is associated with the entailed atom $p$
and the true atoms with an edge from $p$ within $\go$,
$(A^T \cap ext(p))$.
Clearly, this set is sufficient to entail 
and therefore support $p$.
Conversely, the nogood added in \Line{algo:add false} represents that
the atom $p$ cannot be true
along with the true atoms of the current assignment,
or else there would be a contradiction. 

The final section of the algorithm
aims to generate negative entailment nogoods.
For each entailment atom $\beta_{\ont}(p)$
which is not yet assigned false,
we check whether $p$ is within the set of entailed atoms $O$
under the assumption that all atoms in $\KAO \setminus (A^F \cup \{p\})$
are true,
in line \ref{algo:max entailment}.
If $O$ contains neither $p$ nor the contradiction atom $\bot$,
then we can be certain that $p$ cannot be supported via $\ont$ 
whenever all atoms in $A^F \cap ext(\{p\})$ are false.
Therefore, 
in line \ref{algo:add neg nogood} 
we add the corresponding nogood.
Finally, procedure then returns.
If any nogood was added during the process,
\textit{entailed} will be true, 
and otherwise false.

\bigskip
\smallskip
\begin{footnotesize}
    \begin{algorithm}[H]
        \label{algo 3}
        \DontPrintSemicolon
        \SetAlgoVlined
        \begin{multicols}{2}  
            $entailed \leftarrow False$
            
            $\Omega \leftarrow Entailment (A^T \cap \KAO)$
            \label{algo:lower bound}

            \If  {$\bot \in \Omega^+$ \label{algo:if bot}} 
            {
                $\Delta \leftarrow \Delta \cup \gamma_{\ont}^+(\bot, A^T)$
            
                \Return{$(\nabla, True)$}
            }
            \label{algo:end if bot}

            \If{$\Omega^+ \setminus A^T \neq \emptyset$ \label{algo:if true}}
            {
                $entailed \leftarrow True$
            
                \For{$p \in \Omega^+ \setminus A^T$}{
                    $\nabla \leftarrow \nabla \cup \gamma_{\ont}^+(p,A^T \cap ext(\{p\}))$
                    \label{algo:add true}
                }
            }
            
            \columnbreak
        
            \If{$\Omega^- \setminus A^F \neq \emptyset$}
            {
                $entailed \leftarrow True$
            
                \For{$p \in \Omega^- \setminus A^F$}
                {
                    $\nabla \leftarrow \nabla \cup \gamma_{\ont}^+(\bot, A^T\cup \{ p\})$
                    \label{algo:add false}
                }
            }
            
            \For{$p \in \KAO$ such that $\beta_{\ont}(p) \not \in A^F$}
            {
                $O \leftarrow Entailment(\KAO \setminus (A^F \cup \{p\}))$
                \label{algo:max entailment}

                \If{$\bot \not \in O^+$ and $p \not \in O^+$}
                {
                    $entailed \leftarrow True$
            
                    $\nabla \leftarrow \nabla \cup \gamma_{\ont}^-(p, A^F \cap ext(\{p\}))$
                    \label{algo:add neg nogood}
                }
            }
            
            \Return{$(\nabla, entailed)$}

        \end{multicols}
        \caption{\it EntNogoods}
    \end{algorithm}
\end{footnotesize}

\needspace{3\baselineskip}
\begin{example}
    \label{EntNogoods example}
    Below are the results of calling {\it EntNogoods} with 
    different assignments $A$ as input for an HMKNF-KB $\Kintro$
    where $\ont = ((\neg a \vee \neg b) \wedge (\neg a \vee c))$ and $\KAO = \{a, b, c\}$.

    \begin{footnotesize}
    \begin{tabular}{lllllllll}
    \hline
    $A^T$ & $A^F$ & $\Omega^+$ & $\Omega^-$ & $O^+$ &   $p$ &Added Nogood & {\it entailed} & Line \\
    \hline
    $\{a\}$ & $\{\beta_{\ont}(a)\}$ &            &         &            & 
    &&{\it False}&1  \\
    $\{a\}$ & $\{\beta_{\ont}(a)\}$ & $\{a, c\}$ & $\{b\}$ &            & 
    &&{\it False}&2\\
    $\{a\}$ & $\{\beta_{\ont}(a)\}$ & $\{a, c\}$ & $\{b\}$ &            & 
    &&{\it True}&7\\
    $\{a\}$ & $\{\beta_{\ont}(a)\}$ & $\{a, c\}$ & $\{b\}$ &            & 
    $c$& $\gop(c, \{a\})$ &{\it True} &9\\
    $\{a\}$ & $\{\beta_{\ont}(a)\}$ & $\{a, c\}$ & $\{b\}$ &            & 
    $b$& $\gop(\bot, \{a, b\})$ & {\it True} &13\\
    $\{a\}$ & $\{\beta_{\ont}(a)\}$ & $\{a, c\}$ & $\{b\}$ & $\{a, c\}$ & 
    $b$& &{\it True} &15\\
    $\{a\}$ & $\{\beta_{\ont}(a)\}$ & $\{a, c\}$ & $\{b\}$ & $\{\bot\}$ & 
    $c$&& {\it True} &15\\
    $\{a\}$ & $\{\beta_{\ont}(a)\}$ & $\{a, c\}$ & $\{b\}$ & $\{\bot\}$ & 
    && {\it True} &19\\
    \hline
    $\{a, b\}$ & $\emptyset$ &            &         &            & 
    &&{\it False}&1  \\
    $\{a, b\}$ & $\emptyset$ & $\{\bot\}$ & $\emptyset$    &            & 
    &  &{\it False}&2  \\
    $\{a, b\}$ & $\emptyset$ & $\{\bot\}$ & $\emptyset$    &            & 
    & $\gop(\bot, \{a,b\})$&{\it False}&4  \\
    $\{a, b\}$ & $\emptyset$ & $\{\bot\}$ & $\emptyset$    &            & 
    & &{\it True}&5  \\
    \hline
    $\emptyset$ & $\{a, \beta_{\ont}(a)\}$ &            &         &            & 
    &&{\it False}&1  \\
    $\emptyset$ & $\{a, \beta_{\ont}(a)\}$ &  $\emptyset$    & $\emptyset$   &            & 
    &&{\it False}&2  \\
    $\emptyset$ & $\{a, \beta_{\ont}(a)\}$ &  $\emptyset$    & $\emptyset$   & $\{c\}$           & 
    $b$ &&{\it False}&15 \\
    $\emptyset$ & $\{a, \beta_{\ont}(a)\}$ &  $\emptyset$    & $\emptyset$   & $\{c\}$           & 
    $b$ &&{\it True}&17 \\
    $\emptyset$ & $\{a, \beta_{\ont}(a)\}$ &  $\emptyset$    & $\emptyset$   & $\{c\}$           & 
    $b$ & $\gon(b, \{a\})$&{\it True}&18 \\
    $\emptyset$ & $\{a, \beta_{\ont}(a)\}$ &  $\emptyset$    & $\emptyset$   & $\{b\}$           & 
    $c$ & &{\it True}&15 \\
    $\emptyset$ & $\{a, \beta_{\ont}(a)\}$ &  $\emptyset$    & $\emptyset$   & $\{b\}$           & 
    $c$ & $\gon(c, \{a\})$&{\it True}&18 \\
    $\emptyset$ & $\{a, \beta_{\ont}(a)\}$ &  $\emptyset$    & $\emptyset$   & $\{b\}$           & 
    $c$ & &{\it True}&19 \\
    \hline
    \end{tabular}
    \end{footnotesize}
    \smallskip

    Above, the leftmost two columns $A^T$ and $A^F$ display the true and false atoms of the input assignment respectively.
    The next two display the atoms entailed under minimal assumptions as true $\Omega^+$ and false $\Omega^-$.
    Following that is the set of atoms entailed as true under maximal assumptions $O^+$.
    After that is the variable $p$, which reflects the  current value of the iterator 
    within the active for-loop.
    Then, there is a column indicating which, if any, 
    nogood was added to the tracked set in the current step.
    Following which is the Boolean variable {\it entailed} indicating
    whether any nogood has been added.
    Finally, the active line number is displayed on the right.
\end{example}

\section{Related and Future Work}

The first challenge
in applying the theories of this work to the implementation of a conflict-driven solver, 
involves procedures for the generation and management
of unit-resulting entailment and loop nogoods.
Generation of entailment nogoods can follow the approach of \emph{EntNogoods}, 
but requires the integration of ontology reasoners. 
Generating loop nogoods requires additional theoretical work, 
but in principle can be based on detection of unfounded sets.
There must also be an investigation into the ideal policies
for when to attempt generation of these nogoods, 
and when they should be forgotten.
It is essential to keep the set of tracked nogoods small
in order to reduce time spent preforming unit-propagation.
This is likely to be uniquely challenging for HMKNF-KBs.

The second
is the practical challenge of integrating these theories with existing solvers.
The Clingo system 
includes a Theory-enhanced ASP solving API \citep{gebser2016theory},
allowing the user to introduce additional \emph{theory atoms}, 
whose truth is regulated by user-supplied \emph{theory nogoods}.
Using this approach for a tightly integrated framework
such as HMKNF-KBs
is complicated by the support and loop nogoods
already included by Clingo.
In principle
these can be reduced to a subset of our proposed nogoods
by adding supporting rules,
the bodies of whom are theory atoms 
semantically governed entirely by theory nogoods.
However, this approach is non-trivial.
It requires determining both which support rules to add
and the additional theory nogoods required
to construct a set equivalent to our proposed one.

A direct implementation using Clingo's source code 
is also possible but comes with its own complexity.
Another option is to build on the DLVHEX system,
which despite its existing integration with ontology reasoners
is less straightforward to layer our framework upon.
Conversely, a translation plugin that allows DLVHEX to solve HMKNF-KBs as dl-programs 
could provide a useful benchmark.
\comment{
Finally, it should investigated whether the NoHR system
\citep{kasalica2020nohr}
can be taken advantage of in any way.
While it is based on the well-founded rather than 
the stable semantics of HMKNF-KBs it could 
still play a role, for example to help in grounding.
}

\section{Conclusion}

Our work establishes the critical foundation for 
the development of a native conflict-driven solver of HMKNF-KBs. 
We have made significant theoretical contributions 
including the first formulation of a dependency graph,
and first adaption of a completion and loop formulas 
for the formalism. 
These advancements have enabled us to derive a set of nogoods, 
essential for implementing a conflict-driven solver.
We have outlined the architecture of such a solver and 
critically examined both the potential and challenges in 
leveraging existing systems. 
Our findings have significant implications for enhancing the 
efficiency and practicality of reasoning with HMKNF-KBs under the stable model semantics. 
The immediate next steps include the implementation 
of a conflict-driven solver based on our theoretical framework 
and further refinement of the characterizations we have proposed.

\bigskip
\noindent
{\bf Acknowledgements}

\medskip
\noindent
We thank the reviewers for their constructive comments 
that helped improve the presentation of the paper. 
This research was supported in part by Alberta Innovates 
(under grant Alberta Innovates Advance Program 222301990) and by 
the Natural Sciences and Engineering Research Council of Canada 
(under grant NSERC RGPIN-2020-05211). 
For his research, the first author also received 
an NSERC Undergraduate Student Research Award and 
the second author was supported by an Alberta Innovates 
Graduate Student Scholarship.

\comment{
    
\section{Appendix}
\subsection{Saturation}

\begin{theoremEnd}{proposition}
    \label{SAT:rule and satur completion theorem}
    For any \ki\ $\hi$ of an HMKNF-KB $\Kintro$,
    $\hi \models \pro_{rule} \wedge \ont_{satr}(\hi)$
    iff,
    there exists an MKNF interpretation $M$,
    such that $M$ induces $\hi$ 
    and $M \mmodels \pi(\K)$.
\end{theoremEnd}

This theorem presents the problem of whehter a \ki\ $\hi$ of an HMKNF-KB $\K$
is induced by a MKNF interpretation which satisfies $\K$ 
as a satisfiability instance.
Notice that the formula which the \ki\ $\hi$ must satisfy, depends on $\hi$. 
This is different from the completion for ASP,
and is done in order to avoid an exponential blowup in 
the number of formulas.
Unfortunately this means that this version cannot be used directly 
in a SAT solver for the purpose of model discovery, 
however it provides a theoretical basis and intuition for the constraints 
that the nogoods defined in \Section{sec:nogoods}
enforce on particular total assignments.
This same approach will be used in following to define
the support completion and loop formulas of an HMKNF-KB.
}

\end{document}